\documentclass[10pt,twocolumn,letterpaper]{article}
\usepackage[accsupp]{axessibility} 
\usepackage[pagenumbers]{wacv} 

\usepackage{graphicx}
\usepackage{amsmath}
\usepackage{amssymb}
\usepackage{booktabs}
\usepackage{multirow} 
\usepackage[capitalize]{cleveref}
\crefname{section}{Sec.}{Secs.}
\Crefname{section}{Section}{Sections}
\Crefname{table}{Table}{Tables}
\crefname{table}{Tab.}{Tabs.}

\def\wacvPaperID{1915} 
\def\confName{WACV}
\def\confYear{2025}

\begin{document}

\title{ARTIST: Improving the Generation of Text-rich Images with Disentangled Diffusion Models and Large Language Models}

\author{\textbf{Jianyi Zhang}\textsuperscript{1}\thanks{Equal contribution, work was done when
Jianyi Zhang was an intern at Adobe Research. Jianyi Zhang and Yiran Chen disclose
support from grants NSF 2112562 and ARO W911NF-23-2-0224.},
  \textbf{Yufan Zhou}\textsuperscript{2}\footnotemark[1],
  \textbf{Jiuxiang Gu}\textsuperscript{2},
  \textbf{Curtis Wigington}\textsuperscript{2},
  \textbf{Tong Yu}\textsuperscript{2}, \\
  \textbf{Yiran Chen}\textsuperscript{1}\textbf{,}
 \textbf{Tong Sun}\textsuperscript{2} \textbf{,}
 \textbf{Ruiyi Zhang}\textsuperscript{2}\\
  \textsuperscript{1} Duke University,
  \textsuperscript{2} Adobe Research}
\maketitle

\begin{abstract}
Diffusion models have demonstrated exceptional capabilities in generating a broad spectrum of visual content, yet their proficiency in rendering text is still limited: they often generate inaccurate characters or words that fail to blend well with the underlying image. To address these shortcomings, we introduce a novel framework named \textit{{ARTIST}}, which incorporates a dedicated textual diffusion model to focus on the learning of text structures specifically. Initially, we pretrain this textual model to capture the intricacies of text representation. Subsequently, we finetune a visual diffusion model, enabling it to assimilate textual structure information from the pretrained textual model. This disentangled architecture design and training strategy significantly enhance the text rendering ability of the diffusion models for text-rich image generation. Additionally, we leverage the capabilities of pretrained large language models to interpret user intentions better, contributing to improved generation quality. Empirical results on the MARIO-Eval benchmark underscore the effectiveness of the proposed method, showing an improvement of up to 15\% in various metrics. 
\end{abstract}
\section{Introduction}
The field of text-to-image generation has made remarkable progress, especially with the rise of diffusion models \cite{ho2020denoising,rombach2022high,ramesh2022dalle2,saharia2022photorealistic,yang2022diffusion,zhou2023shifted,ruiz2023dreambooth}. These models not only have demonstrated prowess in generating high-fidelity images, but have also showcased versatile applications, such as image inpainting \cite{lugmayr2022repaint,esser2021imagebart,jing2022subspace} , denoising \cite{xie2023diffusion,Yang2023RealWorldDV}, video
generation \cite{han2022card,blattmann2023videoldm}, and style transfer \cite{wang2023stylediffusion,zhang2022inversion}. Along with these advancements, the ability to generate images containing legible text, another realm of significance, is becoming increasingly worthy of attention. Specifically, this need is evident in everyday scenarios, where images with text are commonplace, from advertisements to road signs, posters, and book covers. Crafting these text-rich images manually demands skilled expertise and a significant time commitment. However, a significant shortcoming lies in the current state-of-the-art diffusion-based generative models; they often render text portions that are virtually unreadable, akin to gibberish, undermining the aesthetic and functional value of the generated images. Therefore, if generative AI, powered by diffusion models, can produce such images, it could revolutionize design workflows, inspire creativity, and alleviate designers' workload.
\begin{figure*}[t!]
\vspace{-1cm}
    \centering
    \includegraphics[width=0.75\linewidth]{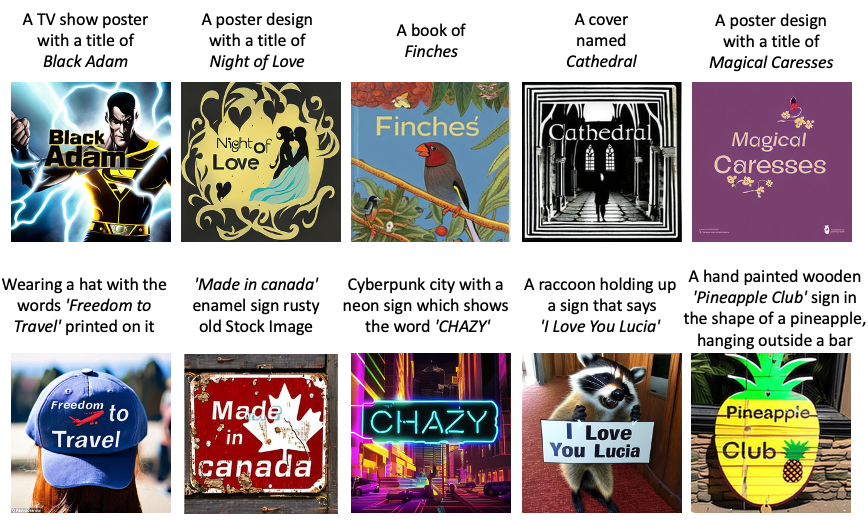}
    \caption{Generated examples from our ARTIST. Our framework adeptly identifies the text intended to be generated in the image from the given prompts, regardless of explicit marking by quotes. The generated text is legible and complements the visual elements, enhancing the overall coherence of the design.}
    \label{fig:examples}
    \vspace{-5mm}
\end{figure*}

To tackle the challenges posed by the quality of text generated by diffusion models, two primary pathways have been explored. Firstly, the traditional approach involves leveraging image-editing tools to superimpose text onto images directly. However, this frequently introduces unnatural artifacts, especially when dealing with intricate textures or varying lighting conditions in the background of the image. In contrast, recent research efforts aim to refine the diffusion models themselves for improved text quality. For example, recent innovations such as Imagen \cite{saharia2022photorealistic}, eDiff-I \cite{balaji2022ediffi}, and DeepFloyd \cite{DeepFloyd} discovered that the use of T5 series text encoders \cite{raffel2020exploring} led to better results compared to using the CLIP text encoder \cite{radford2021learning}. Similarly, Liu et al.\cite{liu2022character} integrated character-aware text encoders to enhance the quality of text rendering. However, these advancements primarily revolve around optimizing text encoders and do not necessarily grant more control over the holistic generation process. On a parallel track, GlyphDraw \cite{ma2023glyphdraw} has enhanced model controllability by conditionally focusing on the positioning and architecture of Chinese characters. Still, its utility remains limited, as it cannot cater to scenarios that require multiple text bounding boxes, making it less suitable for prevalent text-image formats like posters and book covers. TextDiffuser \cite{chen2023textdiffuser}  represents a recent advancement in the realm of enhancing the quality of text in generated images. While it undeniably marks a significant step forward, it is not without limitations. One of the prominent challenges is the model's dependence on manual efforts when it comes to recognizing key terms from prompts, making the process less efficient than desired. Additionally, even though TextDiffuser exhibits improved capabilities, the accuracy rate of its Optical Character Recognition (OCR) evaluation on generated text still leaves room for optimization.

In this paper, our primary objective is to develop a more efficient system for text-to-image generation. This system would eliminate the need for manual post-production adjustments, ensuring superior text quality within generated images. Upon meticulous investigation, we have identified the primary challenges we aim to address. Firstly, automation of accurately discerning which words from a provided text prompt should be incorporated into the image. In platforms such as TextDiffuser, this identification requires human involvement, typically by highlighting specific terms with quotation marks, diminishing the platform's efficiency and automation. Second, there is the challenge of adeptly generating images that seamlessly integrate top-quality text, ensuring adherence to a predetermined layout, and emphasizing in the generated images.

\begin{figure*}[t]
\vspace{-1cm}
    \centering
    \includegraphics[width=0.85\linewidth]{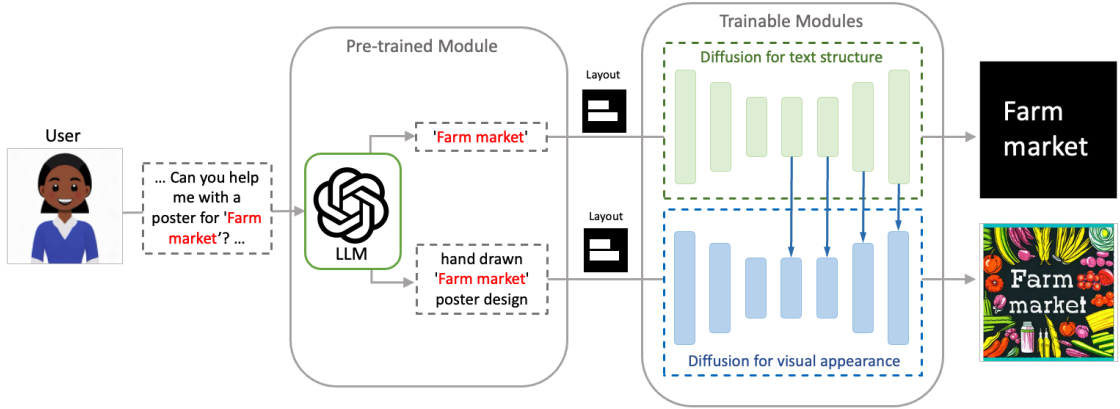}
    \vspace{-0.1in}
    \caption{Illustration of the proposed ARTIST framework. A large-language model (LLM) is utilized to analyze the user's intention. Two diffusion models will be trained to learn text structure and other visual appearance respectively. Given a user input, the LLM will output keywords, layout and text prompts, which will be fed into our trainable modules to generate target images.}
    \label{fig:framework}
    \vspace{-0.1in}
\end{figure*}

To address the aforementioned challenges, we turned to the latest advancements in natural language processing and large language models (LLMs)~\cite{OpenAI2023GPT4TR,touvron2023llama,wang2024coreinfer,joren2024sufficient,yao2024federated,zhang2022join,kuo2023dacbert,zhang2023reaugkd}. LLMs have showcased outstanding expertise in understanding and processing intricate linguistic patterns, making them perfectly suited for our needs. Inspired by their prowess, we formulated a strategy that employs large language models to surmount the initial hurdle of pinpointing the keywords. To tackle the subsequent challenge, we introduce a novel two-stage approach named ARTIST, detailed in Figure \ref{fig:framework}. Specifically, ARTIST's dual stages are dedicated to mastering text structure and refining visual aesthetics in that order. The moniker ``ARTIST" encapsulates the essence of \textit{the Ability of Rendering Text can be Improved by diSentanglemenT}. By merging our proposed methodology with LLMs' capabilities, we have achieved a marked improvement in the quality of text embedded within images.

Our contributions can be summarized as follows.
\begin{itemize}
    \item We proposed an efficient framework that improves the controllability and quality of text generated within images generated by diffusion models.\vspace{-2mm}
    \item Through combining LLMs and diffusion models, we leveraged the capabilities of LLMs like GPT-4 \cite{OpenAI2023GPT4TR} to devise an efficient prompt for understanding complex, even open-domain, user instructions.\vspace{-2mm}
    \item We pioneered the first training strategy that separates learning text structure and visual appearance, boosting 
    performance on existing text rendering benchmarks by up to 15\% in terms of text OCR accuracy.\vspace{-2mm}
\end{itemize}

\section{Related Work}
\subsection{Text-to-image Generation}
Recent advances in text-to-image generation models can be categorized into two primary categories: autoregressive frameworks \cite{ramesh2021zero,yu2022scaling} and diffusion-based models \cite{ramesh2022dalle2,saharia2022photorealistic}, have been substantial. The latter, particularly diffusion-centric models, have gained significant momentum in recent times. The crux of these models lies in the conversion of textual prompts into latent representations, subsequently relying on diffusion mechanisms to formulate images. Several renowned models, such as Stable Diffusion \cite{rombach2022high}, DALL-E 2 \cite{ramesh2022dalle2} and Imagen \cite{saharia2022photorealistic} have set benchmarks in this domain. The challenge of controlled generation remains at the forefront. Contemporary image generation models, with an emphasis on diffusion, heavily prioritize text-based guidance, making it a promising vector for refining their controllability. The inherent complexities of gaining exact control through textual means are evident. For example, ControlNet \cite{zhang2023adding} provides an architectural blueprint that adapts pretrained diffusion models to accommodate a variety of input conditions. Although offering a higher level of flexibility, obtaining such condition signals often necessitates manual intervention and might be restrictive for broader conceptual applications. Methods such as GLIGEN \cite{li2023gligen} advocate for open-set image generation by employing grounding tokens to define the spatial parameters of objects. We agree with this approach, integrating elements of the grounding token methodology. 

\subsection{Text Rendering \& OCR}

Despite the rapid development of diffusion-based Text-to-image Generation, existing methodologies still struggle with generating precise and consistent textual renderings. Various approaches, like Imagen \cite{saharia2022photorealistic}, eDiff-I \cite{balaji2022ediffi}, and DeepFolyd \cite{DeepFloyd}, have leveraged the prowess of expansive language models (notably large T5 \cite{raffel2020exploring}) to bolster their textual accuracy. The study in \cite{liu2022character} highlights a limitation in which traditional text encoders overlook token length, prompting them to propose a character-sensitive alternative. Simultaneously, GlyphDraw \cite{ma2023glyphdraw} focuses on producing superior images integrated with Chinese texts, guided by textual positioning and glyph imagery. GlyphControl \cite{yang2024glyphcontrol} further enhances this approach by adjusting the text alignment according to its location, implicitly incorporating elements like font size and text box positioning. A recent study, AnyText \cite{tuo2023anytext}, utilizes a diffusion pipeline with an auxiliary latent module and a text embedding module to improve the text generation, editing, and integration with the image background. The techniques of Textdiffuser \cite{chen2023textdiffuser} harnesses the Transformer model to discern keyword layouts, promoting multiline text generation. Then, it further employs character-based segmentation masks as a prior, offering flexibility in control to cater to user specifications. However, it still depends on manual efforts when it comes to recognizing key terms from prompts. Although Textdiffuser-2 \cite{chen2023textdiffuser-2} employs LLMs to enhance the interpretation of prompts, the improvements in the quality of generated images with text remain modest. This indicates that although the system has improved in understanding user inputs, converting these advancements into better visual results still demands additional optimization. Hence, we believe it is necessary to develop a more effective framework that captures the intricacies of text representation and seamlessly integrates textual structures into images, aiming for more coherent outputs. 
\vspace{-5mm}
\paragraph{OCR} Optical Character Recognition (OCR) is a long-established academic endeavor \cite{white1983image,cash1987optical}. In the last decade, this field has seen remarkable progress, impacting a range of applications including recognition of car license plates \cite{Iraqi_car_license}, autonomous vehicle navigation \cite{Detecting_symbols6957755,wu2023ocr}, and its incorporation into foundational models such as GPT \cite{huang2023language,shen2023hugginggpt}. In this paper, OCR serves as a pivotal evaluation metric used to critically assess the quality of the generated text and provide a comprehensive understanding of our model's performance in realistic scenarios.

\section{Methodology}
As revealed in previous works \cite{ramesh2022dalle2,saharia2022imagen,chen2023textdiffuser}, generating an image with text rendered on it is still challenging. We suspect that this happens because of two major reasons: 
\vspace{-2mm}
\begin{itemize}
    \item It is challenging to simultaneously learn visual appearance and text structure with single model;\vspace{-2mm}
    \item Existing datasets are not able to cover all the words and their possible combinations, making it hard to learn text structure from these limited noisy data.
\end{itemize}
\vspace{-2mm}
In this work, we propose to mitigate the aforementioned challenges by utilizing separate modules to learn text structure and visual appearance. Furthermore, these two modules are trained separately, making it possible to learn text structure with synthetic data constructed by ourselves, which also tackles the problem of limited data. To ensure disentanglement and prevent information leakage, our text module and visual module take different prompts as inputs. However, it can be inefficient and user-hostile if these prompts have to be manually designed by the user themselves at inference. Fortunately, because of the recent success of large-language models (LLMs)~\cite{brown2020GPT3,touvron2023llama2, zhang2024towards,zhang2024sled,zhang2024min,zhang2024mllm}, we propose to utilize pre-trained LLMs to infer user's intention, provide accurate prompts for both modules. With the help of LLM, user's input can be either precise or vague, leading to a better interactive experience. Due to the disentangled nature of our framework, it is also friendly for future implementations that leverage distributed training \cite{fedcbs,yang2020federated,du2022rethinking,tang2022fade,jia2025unlocking,hao2021towards,zhang2022next,yang2021flop} to utilize computational resources while safeguarding data privacy.

Our proposed framework is termed ARTIST, because it illustrates that \textit{the Ability of Rendering Text can be Improved by diSentanglemenT}.
Our proposed framework is illustrated in Figure \ref{fig:framework}, with details discussed below. As we show in Section \ref{sec: experiments}, our proposed framework outperforms the previous state-of-the-art (SOTA) in terms of image fidelity, image-prompt alignment, and accuracy of generated texts.

\subsection{LLM-based Prompt Understanding} \label{sec:LLM_PU}
TextDiffuser~\cite{chen2023textdiffuser} proposes to train a transformer model to extract texts that are expected to be shown on generated images. However, there appears to be a limitation associated with this method: their model is only capable of detecting words enclosed by quotation marks. This happens because most keywords from the training dataset are enclosed by quotation marks inside the captions. A model trained on these samples will lack generalization ability or even be overfitting. Moreover, their model even fails to generate the desired image according to the prompt ``a poster of Batman with the word Batman on it", as it does not understand that the word ``Batman" should be presented on the image.

On the contrary,
we propose to provide pre-trained LLM with an open-domain prompt and rely on its capability to autonomously identify the essentials. Because of its vast open-domain knowledge, LLM is able to better understand user intents and thus can generalize to complicated scenarios. Given vague prompts, LLM is able to discern and propose which words or text elements to incorporate, resulting in more coherent and aesthetically appealing suggestions.

\subsection{Learning Text Structure}
\label{sec: learntext}
A diffusion model, denoted as a text module, is introduced to learn text structures. Specifically, this text module is trained to take bounding boxes as input and generates a black-white image with only text on it. From Bayesian learning \cite{zhang2020variance,zhang2020stochastic,zhang2019cyclical,zhao2018selfadversarially,chen2022we} perspective, our text module essentially learns a more reliable prior distribution for the subsequent generation process.
Because we do not require it to learn any visual effects, the prompt for this module is the word to be rendered and the dataset can be simply constructed using standard rendering libraries
Specifically, we construct two large-scale datasets to train this module, as described below.
\begin{figure}[t]
    \centering
    \includegraphics[width=0.3\textwidth]{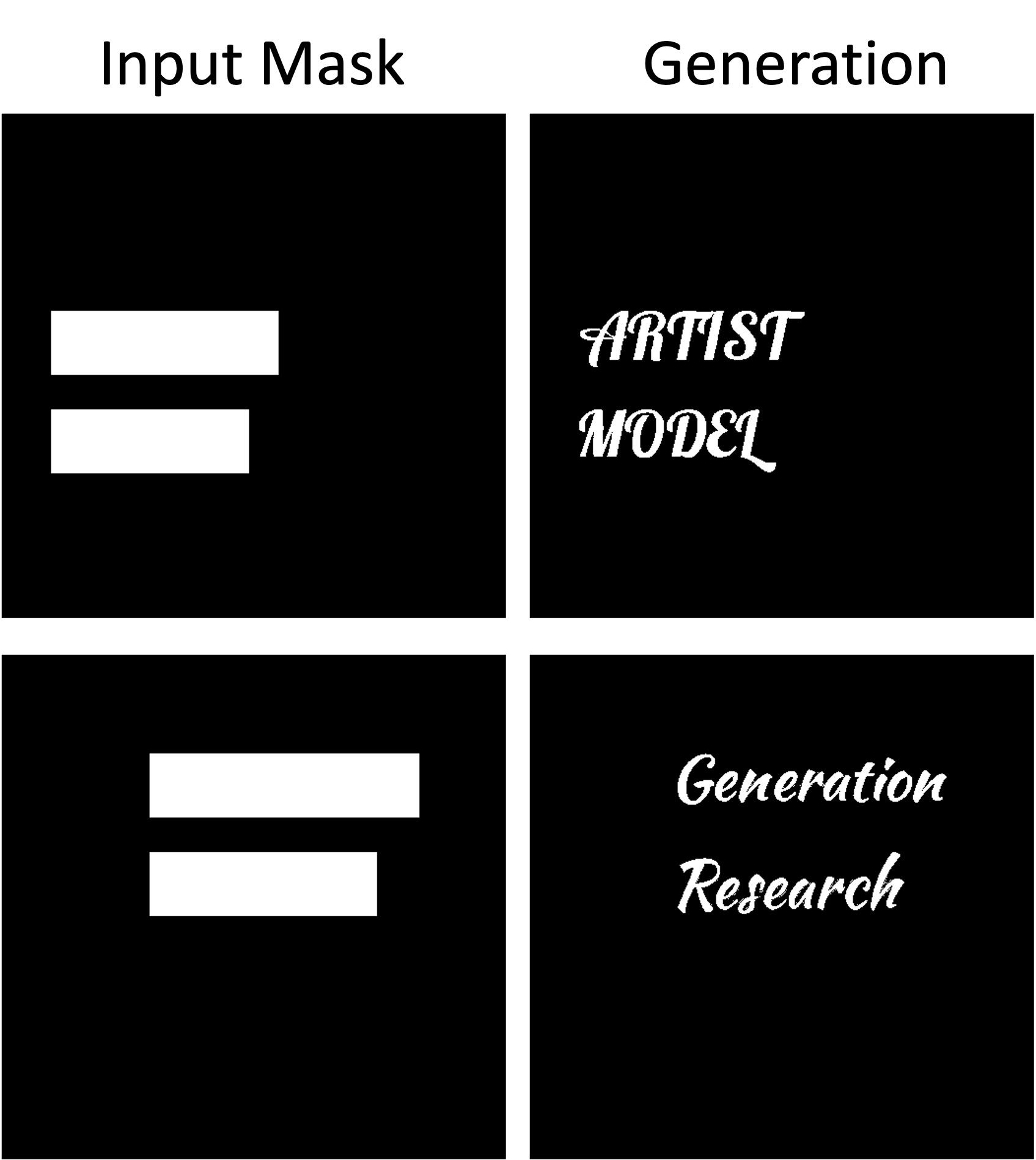}
     \caption{Generated examples from our text module, along with input masks.}\label{fig:text_module}
     \vspace{-0.5cm}
\end{figure}
\paragraph {Word-level dataset:} Our first dataset consists of around 10 millions of black-white images with only single word on them. Specifically, to construct each data sample, we first randomly select a word from the vocabulary of CLIP text encoder, then render this word with random font and size on a black image. Meanwhile, we can also obtain the ground-truth bounding box for each rendered word. This dataset will be further augmented during training by randomly moving both the word and bounding box to a new position, resulting in more effective samples;\vspace{-2mm}
  \paragraph {Sentence-level dataset:} Although our word-level dataset contains massive structure information of English words, it contains no layout information about how should these words be placed and combined on the target image. To this end, we construct our second dataset which contains 50 millions of black-white images by utilizing MARIO-10M dataset~\cite{chen2023textdiffuser}. Specifically, we use the ground-truth text and layout information from MARIO-10M samples, and render the same text with randomly selected fonts on black images following the same layout.

After constructing both datasets, we train the text module in two stages, inside the latent space of a VAE~\cite{kingma2013auto} following Rombach et al.~\cite{rombach2022high}. In the first stage, the diffusion model is trained to take a single bounding box and a target word as inputs, then generate a black-white image with the word rendered on it. Then, this module will be fine-tuned on our sentence-level dataset as the second stage so that it can take multiple bounding boxes and multiple s as inputs. 
Our s are denoted as \( \mathcal{P} = \{p_1, p_2, \ldots, p_n\} \), composed of the words to be generated in the image. Our input mask is an image containing the corresponding bounding boxes \( \{m_1, m_2, \ldots m_n\} \) that indicate the position of each word. 
During training, the original text-only image and input mask image are first encoded into latent space features $\mathbf{z}$ and $\mathbf{m}$. Then we sample a time step $t \sim$ Uniform $\left(0, T\right)$ and a Gaussian noise $\boldsymbol{\epsilon}$ to corrupt the original feature, yielding $\mathbf{z}_t =\sqrt{\bar{\alpha}_t} \mathbf{z}+\sqrt{1-\bar{\alpha}_t} \boldsymbol{\epsilon}$ where $\overline{\alpha}_T$ is the coefficient of the diffusion process introduced in \cite{ho2020denoising}. $\mathbf{z}_t$ and $\mathbf{m}$ are concatenated in the feature channel as input for diffusion model, which will be trained with the diffusion loss between the sampled $\boldsymbol{\epsilon}$ and the predicated noise $\boldsymbol{\epsilon}_\theta$:
\begin{align} 
    \mathcal{L}_{\text{text}}= \mathbb{E}\left[\left\|\boldsymbol{\epsilon}-\boldsymbol{\epsilon}_\theta\left(\mathbf{z}_t, \mathbf{m}, \mathcal{P}, t \right)\right\|_2^2\right].
    \vspace{-2mm}
\end{align}
Some generated examples are shown in Figure \ref{fig:text_module}, which illustrates that our text module can successfully generate a target image with desired texts on it in different styles. In our implementation, MARIO-10M dataset~\cite{chen2023textdiffuser} is used to train this module following previous work for a fair comparison. $\overline{\mathcal{P}}, \mathcal{P}, \mathbf{m}$ have already been prepared in MARIO-10M. At inference time, LLM will be utilized to infer $\overline{\mathcal{P}}, \mathcal{P}, \mathbf{m}$ as mentioned in Section \ref{sec:LLM_PU}. 

\subsection{Learning Visual Appearance}
After training the text module, we would like to utilize its learned knowledge to generate high-fidelity images containing text. To this end, we propose to inject intermediate features from our text module into our visual module, which is also a diffusion model.
For each intermediate feature from the mid-block and up-block layers of text module, we propose to use a trainable convolutional layer
to project the feature and add it element-wisely onto the corresponding intermediate output feature of the visual module. We have also tested different architectures, the comparison will be provided in later experiment section. 

During the training of the visual module, the text module will be frozen, and only the newly introduced layers and visual module will be fine-tuned.
Differently from the text module described in Section \ref{sec: learntext}, whose training text only contains the target words to be rendered, the training data for our visual module has to contain visual descriptions of the image so that the model can successfully learn to generate visual contents based on the user's input. 

Let $\overline{\mathcal{P}}$ be the prompt which contains visual descriptions of the image, $\mathcal{P}$ be the s as defined in Section \ref{sec: learntext}.
We feed $\mathcal{P}$ and the input mask $\mathbf{m}$ as mentioned in Section \ref{sec: learntext} into the pre-trained text module. 
The intermediate features from the text module denoted as $\{f_i(\mathbf{z}_t, \mathbf{m}, \mathcal{P}, t)\}_{i=1}^k$ will be injected into the visual module, where $k$ stands for the number of intermediate features. The visual module will be trained with diffusion loss:
\begin{align}
\vspace{-4mm}
    \mathcal{L}_{\text{visual}} = \mathbb{E}\left[ \left\| \epsilon - \epsilon_\theta\left(x_t, \{f_i(z_t, m, \mathcal{P}, t)\}_{i=1}^k, m, \overline{\mathcal{P}}, t\right)\right\|_2^2 \right] \nonumber \vspace{-6mm}
\end{align}
where $\mathbf{x}_t=\sqrt{\bar{\alpha}_t} \mathbf{x}+\sqrt{1-\bar{\alpha}_t} \boldsymbol{\epsilon}$ denotes the corrupted VAE feature of ground-truth image.

\section{Experiments}\label{sec: experiments}
\begin{table*}[t]
    \vspace{-0.5cm}
    \centering
    \caption{Results on MARIO-Eval benchmark, our ARTIST outperforms previous methods.}
    \label{tab:main_results}
    \scalebox{0.75}{
        \begin{tabular}{cccccccc}
        \toprule
        Metrics & SD & Fine-tuned SD & ControlNet & DeepFloyd & TextDiffuser & ARTIST-TD & ARTIST\\
        \midrule
            FID $(\downarrow)$ & 51.295 & \textbf{28.761} & 51.485 & 34.902 & 38.758 &  36.579 & 38.43\\
            CLIP Score $(\uparrow)$ & 0.3015 & 0.3412 & 0.3424 & 0.3267 & 0.3436 & 0.3466& \textbf{0.3482}\\
            OCR Accuracy $(\uparrow)$ & 0.0178 & 0.0154 & 0.2705 & 0.0457 & 0.5712 & {0.6298} & \textbf{0.7373} \\
            OCR Precision $(\uparrow)$ & 0.0192 & 0.1777 &0.5391 & 0.1738 & 0.7795 & 0.8237 & \textbf{0.8681} \\
            OCR Recall $(\uparrow)$ & 0.0260 & 0.2330 & 0.6438 & 0.2235 & 0.7498 & {0.7986} & \textbf{0.8677} \\
            OCR F-measure $(\uparrow)$ & 0.0221 & 0.2016 & 0.5868 & 0.1955 & 0.7643 & {0.8110} & \textbf{0.8679}\\
        \bottomrule
        \end{tabular}
    }
\end{table*}

\begin{figure*}[t!]
    \centering
    \includegraphics[width=0.75\linewidth]{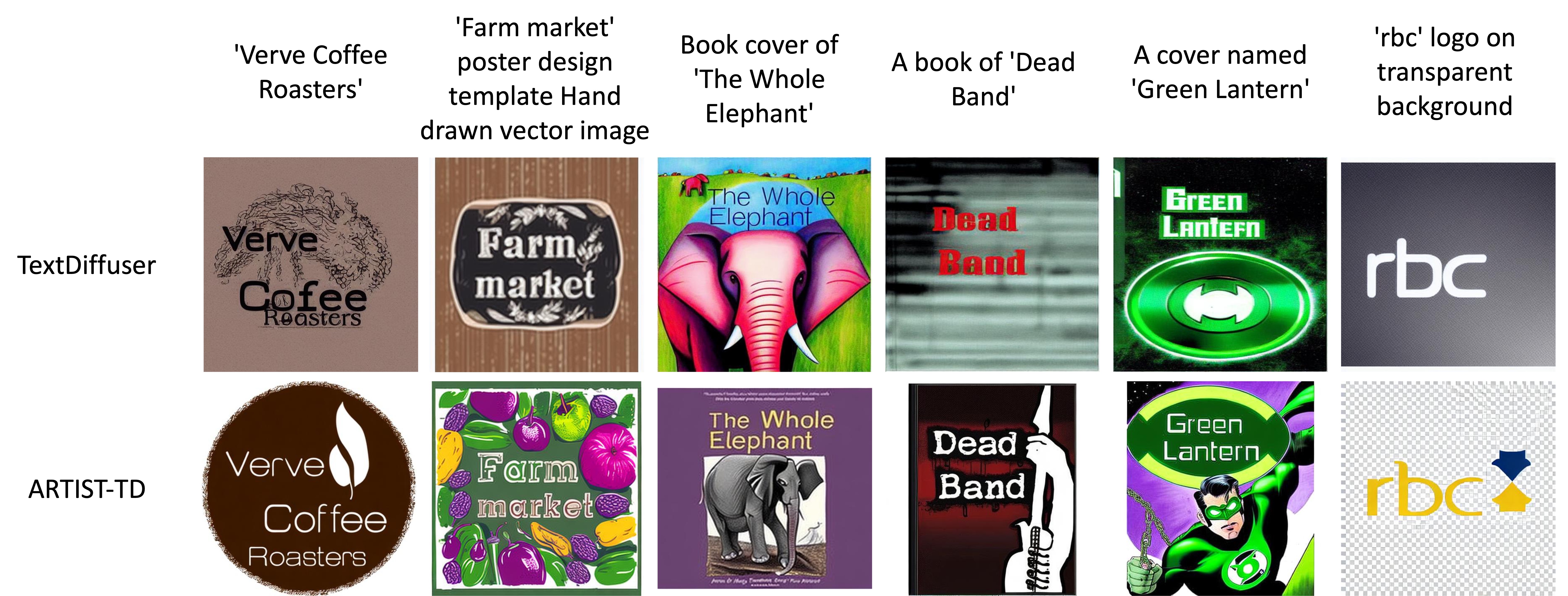}
    \caption{Comparison with TextDiffuser on MARIO-Eval benchmark. Layout generated by TextDiffuser is used as input conditions for both models for fair comparison. }
    \label{fig:comparison}
\end{figure*}

\subsection{Implementation Details}
Our experiments are conducted on 8 Nvidia A100 GPUs, with Hugging Face Diffusers~\cite{von-platen-etal-2022-diffusers}. Both text and visual modules are initialized from the pre-trained Stable Diffusion \cite{rombach2021LDM} checkpoint. The text module is first pre-trained on our word-level synthetic dataset for 400,000 steps, then further fine-tuned for 200,000 steps on our sentence-level synthetic dataset. With the frozen text module, our visual module is trained for 250,000 steps on MARIO-10M dataset \cite{chen2023textdiffuser}. AdamW~\cite{loshchilov2017AdamW} optimizer is used in all training stages, with a learning rate of 1e-5 and a weight decay of 1e-2. The batch size is set to 128.

We first compare our proposed framework with previous methods, including Stable Diffusion~\cite{rombach2021LDM} (denoted as SD), ControlNet~\cite{zhang2023controlnet}, DeepFloyd~\cite{DeepFloyd} and TextDiffuser~\cite{chen2023textdiffuser} in terms of common metrics such as OCR accuracy and FID.  We also finetune a Stable Diffusion on MARIO-10M dataset for a more comprehensive comparison. Then, we directly compare our method with the latest approaches like AnyText \cite{tuo2023anytext} and TextDiffuser 2 \cite{chen2023textdiffuser-2} through human evaluation, as it encapsulates the capabilities of the above traditional metrics while also assessing subjective aspects crucial for user experience and practical application effectiveness. Two variants of our proposed framework are evaluated, which are denoted as ARTIST-TD and ARTIST, respectively, indicating whether LLM is utilized. Specifically, ARTIST-TD directly uses the pretrained transformer from TextDiffuser instead of LLM to obtain bounding boxes and keywords based on input prompts. Thus, comparing ARTIST-TD with TextDiffuser can straightforwardly show the effectiveness of our training strategy, as they share exactly the same layout and keyword conditions.

Note that although there are two separate diffusion models in our framework, our computation requirement is still similar to the previous SOTA TextDiffuser. This is because TextDiffuser also utilizes an extra U-Net, which is designed for character-aware loss as a regularization term.

\subsection{Main Results}

\paragraph{MARIO-Eval benchmark} To start with, we conduct experiments on MARIO-Eval benchmark proposed in ~\cite{chen2023textdiffuser}, which contains 5,414 prompts in total. 4 images are generated for each prompt to compute the CLIP score~\cite{radford2021CLIP}, Fréchet Inception Distance (FID)~\cite{heusel2017FID} and OCR evaluations. Specifically, CLIP score is obtained by calculating the cosine similarity between generated images and prompts by using the features extracted with pre-trained ViT/B-32 CLIP model. OCR evaluation is performed with MaskSpotterv3~\cite{liao2020maskspotter} following \cite{chen2023textdiffuser}. The main results are presented in Table~\ref{tab:main_results}, from which we can see that our ARTIST outperforms the previous SOTA TextDiffuser in all metrics, even without the help of LLM. Some generated examples are provided in Figure \ref{fig:examples}, and more results are provided in the Appendix because of the limited space. 

For a more straightforward comparison, we also provide some generated examples in Figure~\ref{fig:comparison}. Although both TextDiffuser and ARTIST-TD are given the same layout and keyword conditions, we can see that ARTIST-TD generates images with better harmonization between painted text and background. ARTIST-TD is also able to generate many visual features that correspond to keywords such as ``Green Lantern'' and ``transparent'' in the prompt. This is because the learning of text structure and visual appearance is better disentangled, which leads to more efficient learning of both aspects compared to TextDiffuser.

\vspace{-3mm}
\paragraph{ARTIST-Eval benchmark} As the reader may notice from Figure \ref{fig:examples} and \ref{fig:comparison}, most keywords in MARIO-Eval prompts are enclosed by quotation marks. In other words, a model can easily obtain promising accuracy in keyword prediction even if it is over-fitting and simply extracts words enclosed by quotation marks. Thus MARIO-Eval benchmark is not a reasonable benchmark to justify model performance on open-domain instructions. To this end, we propose a new ARTIST-Eval benchmark, which contains 500 pairs of prompts and keywords. Details about constructing the benchmark and some examples are provided in the Appendix \ref{app:artist_benchmark}. 
The keywords in ARTIST-Eval benchmark, which are expected to be shown in the resulting images, 
are designed to be contained in the prompts so that we can compare all the methods fairly.
However, keywords are not always enclosed by quotation marks. Similar to MARIO-Eval benchmark, 4 images are generated for each prompt to calculate CLIP score and OCR evaluations. The results on ARTIST-Eval benchmark are presented in Table \ref{tab:artist_benchmark}. From the results we can see that both ARTIST-TD and ARTIST outperform related methods, while ARTIST obtains a huge performance boost, indicating the benefits of using LLM.
\begin{table*}[t!] 
\vspace{-0.5cm}
    \centering
    \caption{Results on ARTIST benchmark, our proposed framework outperforms all previous methods.}
    \label{tab:artist_benchmark}
    \scalebox{0.8}{
        \begin{tabular}{cccccccc}
        \toprule
        Metrics & SD & Fine-tuned SD & ControlNet & DeepFloyd & TextDiffuser & ARTIST-TD & ARTIST \\
        \midrule
            CLIP Score $(\uparrow)$ & 0.3387 & 0.3440 & 0.3105 & 0.3419 & 0.3100 & 0.3225 & \textbf{0.3545}\\
            OCR Accuracy $(\uparrow)$ & 0.0065 &  0.0300 &  0.0830 & 0.0545 & 0.1345 & 0.2185 & \textbf{0.6530} \\
            OCR Precision $(\uparrow)$ & 0.1031 &  0.1931 & 0.1885 & 0.2225 & 0.2160 & 0.2850 & \textbf{0.8166} \\
            OCR Recall $(\uparrow)$ &  0.1213 &  0.2352 &  0.2304  & 0.2669 & 0.1994 &  0.2779 & \textbf{0.8090} \\
            OCR F-measure $(\uparrow)$ & 0.1114 &  0.2121 & 0.2074 &  0.2427 & 0.2073 & 0.2814 & \textbf{0.8128} \\
        \bottomrule
        \end{tabular}
    }
\end{table*}
\begin{figure*}[t!]
    \centering
    \includegraphics[width=0.75\linewidth]{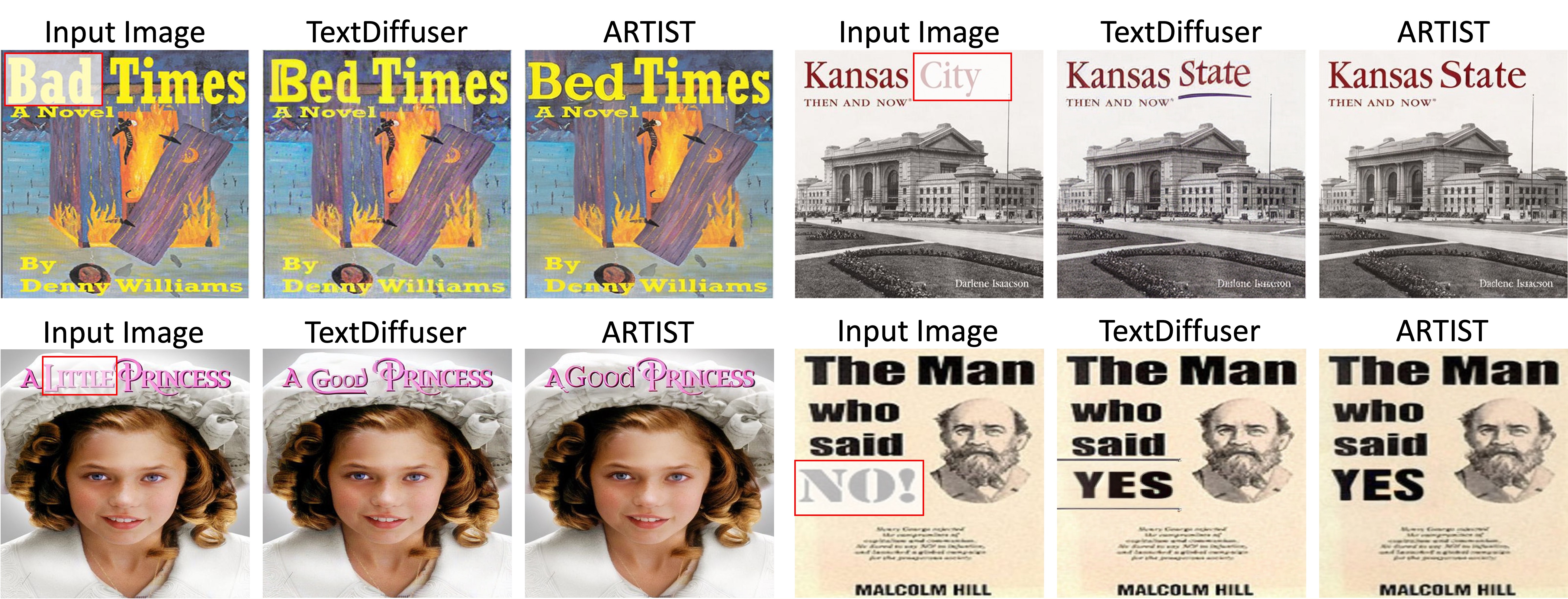}
    \caption{Generated examples in inpainting task, where the masked regions are indicated by red rectangles. Prompts used are ``a book cover of Bed Times", ``a book cover for Kansas State", ``a poster for A Good Princess" and ``a poster for The Man Who Said YES".}
    \label{fig:inpainting}
    \vspace{-0.1in}    
\end{figure*}
\vspace{-3mm}
\paragraph{Keywords Identification} To quantitatively evaluate the contribution of LLM, we conduct experiments of keywords identification. Different models are asked to extract keywords from prompts of ARTIST-Eval benchmark.
The results are shown in Table \ref{tab:keywords_identification}, where we can conclude that LLM indeed leads to huge improvement in detecting keywords.
\begin{table}[ht]
    \centering
    \caption{Large-language models improves keywords identification by a large margin. }
    \label{tab:keywords_identification}    
    \scalebox{0.9}{
        \begin{tabular}{ccccc}
        \toprule
         \multirow{2}{*}{Models} & \multicolumn{4}{c}{Keywords Identification Evaluation}\\
         & Acc $(\uparrow)$ & P $(\uparrow)$ & R $(\uparrow)$ & F1 $(\uparrow)$ \\
        \midrule
            TextDiffuser & 0.6320 & 0.6412  & 0.6397 & 0.6404 \\
            GPT-3.5-Turbo & 0.8300 & 0.9582 & 0.9496 & 0.9539 \\
            GPT-4 & 0.9380 & 0.9729 & 0.9893 & 0.9810 \\
        \bottomrule
        \end{tabular}
    }
\end{table}

\paragraph{Image inpainting}
Similar to TextDiffuser, the proposed ARTIST can also be trained to perform an image-inpainting task. We trained another ARTIST following the setting from Chen et al.~\cite{chen2023textdiffuser} so that the model is asked to perform image inpainting instead of image generation with a probability of 0.5 during training. The resulting ARTIST is able to perform both tasks at inference time. Some generated examples are provided in Figure \ref{fig:inpainting}, along with the comparison between ARTIST and TextDiffuser. Similarly to the whole-image generation task, our ARTIST leads to more accurately rendered texts and more harmonized images.


\begin{table}[ht]
    \centering
    \caption{Intermediate features from text module's U-Net decoder lead to better performance.}
    \label{tab:encoder_decoder}    
    \scalebox{0.9}{
        \begin{tabular}{cccc}
        \toprule
         & U-Net Encoder & U-Net Decoder \\
        \midrule
        FID $(\downarrow)$ & 37.454 & \textbf{36.579} \\
        CLIP Score $(\uparrow)$ & 0.3455 & \textbf{0.3466} \\
        \midrule
        \multicolumn{3}{l}{OCR Evaluation} \\
        \midrule
        Accuracy $(\uparrow)$ & 0.4935 & \textbf{0.6298} \\
        Precision $(\uparrow)$ & 0.7312 & \textbf{0.8237} \\
        Recall $(\uparrow)$ & 0.7293 & \textbf{0.7986} \\
        F-measure $(\uparrow)$ & 0.7302 & \textbf{0.8110} \\
        \bottomrule
        \end{tabular}
    }
\end{table}

\subsection{Human Evaluation}
\label{sec:humaneval}
In the section, we compare our method with GlyphControl \cite{yang2024glyphcontrol}, AnyText \cite{tuo2023anytext}, Controlnet \cite{zhang2023controlnet}, Textdiffuser \cite{chen2023textdiffuser}, and Textdiffuser-2 \cite{chen2023textdiffuser-2} in terms of human evalution in Table \ref{tab:models_comparison}. We referred to Textdiffuser's survey design for our evaluation. We collected 49 cases, which is three times the number of cases in Textdiffuser's study. In each case, we provided images generated by the aforementioned methods to the participants and asked them to rate them based on text rendering quality and image-text matching. The rating scores ranged from 1 to 4, with 4 indicating the best. Similar to Textdiffuser's study, We collected 30 survey responses in total. (The prompts for these 49 cases were sampled from various datasets. We included 7 from ARTIST-Eval, 7 from ChineseDrawText, 2 from DrawBenchText, 7 from DrawTextCreative, 12 from LAIONEval4000, 7 from OpenLibraryEval500, and 7 from TMDBEval500, considering their different dataset sizes.) We provide some examples from the survey in the appendix. 
\begin{table*}[t]

\centering
\caption{Comparison of different models in text rendering and image-text matching.}
\scalebox{0.9}{
\begin{tabular}{lcccccc}
\hline
\textbf{Metric} & \textbf{GlyphControl} & \textbf{AnyText} & \textbf{Controlnet} & \textbf{Textdiffuser} & \textbf{Textdiffuser-2} & \textbf{Ours} \\
\hline
Text Rendering Quality & 2438 & 3333 & 2402 & 2510 & 2922 & 4736 \\
Image-Text Matching & 3009 & 3898 & 2740 & 2824 & 3115 & 4398 \\
\hline
\end{tabular}
}
\label{tab:models_comparison}
\end{table*}
\begin{figure*}[ht]
    \centering
    \includegraphics[width=0.95\linewidth]{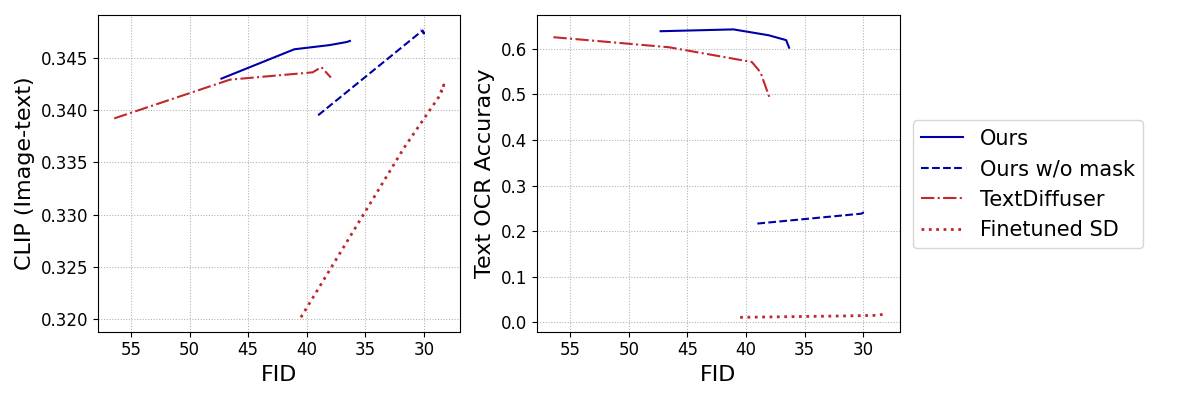}
    \caption{Ablation study of comparing proposed ARTIST with baseline models. The proposed framework is able to generate more accurate text and obtain better image-text similarity across different classifier-free guidance scales.}
    \label{fig:ablation_study}
    \vspace{-4mm}
\end{figure*}

\subsection{Ablation Studies}
\paragraph{Robustness} Although character segmentation mask is used in our implementation following ~\cite{chen2023textdiffuser}, the proposed ARTIST should lead to improvements no matter whether character segmentation mask is used or not. To verify this claim, we conduct an ablation study to show the robustness of ARTIST. Specifically, we trained an ARTIST model without character-level segmentation mask, which is to be compared with the fine-tuned Stable Diffusion model. For all the models, we apply 50 sampling steps and classifier-free guidance~\cite{ho2022classifier-freeguidance} of $\omega \in \{2.5, 5, 7.5, 10, 15\}$ during inference. 
In general, a larger $\omega$ leads to better FID and CLIP scores, while it may lead to a worse OCR evaluation. The comparison is provided in Figure~\ref{fig:ablation_study}, from which we can conclude that our ARTIST leads to improvement across different settings because it leads to better alignment of the image and the text, image fidelity, and OCR accuracy compared to baselines. 

\paragraph{Network architecture} We also conduct an experiment in which a ControlNet-like architecture is used to introduce the learned features into the visual module. Specifically, intermediate features of down-blocks and mid-blocks from the text module's U-Net are projected into the visual module. The experiment is conducted on MARIO-Eval benchmark so that FID, CLIP score, OCR results can all be reported. LLM is not used in this part. The comparison is provided in Table \ref{tab:encoder_decoder}, from which we can see that the intermediate features from the text module's U-Net decoder lead to better performance.

\begin{figure}[ht]
    \centering
    \includegraphics[width=0.95\linewidth]{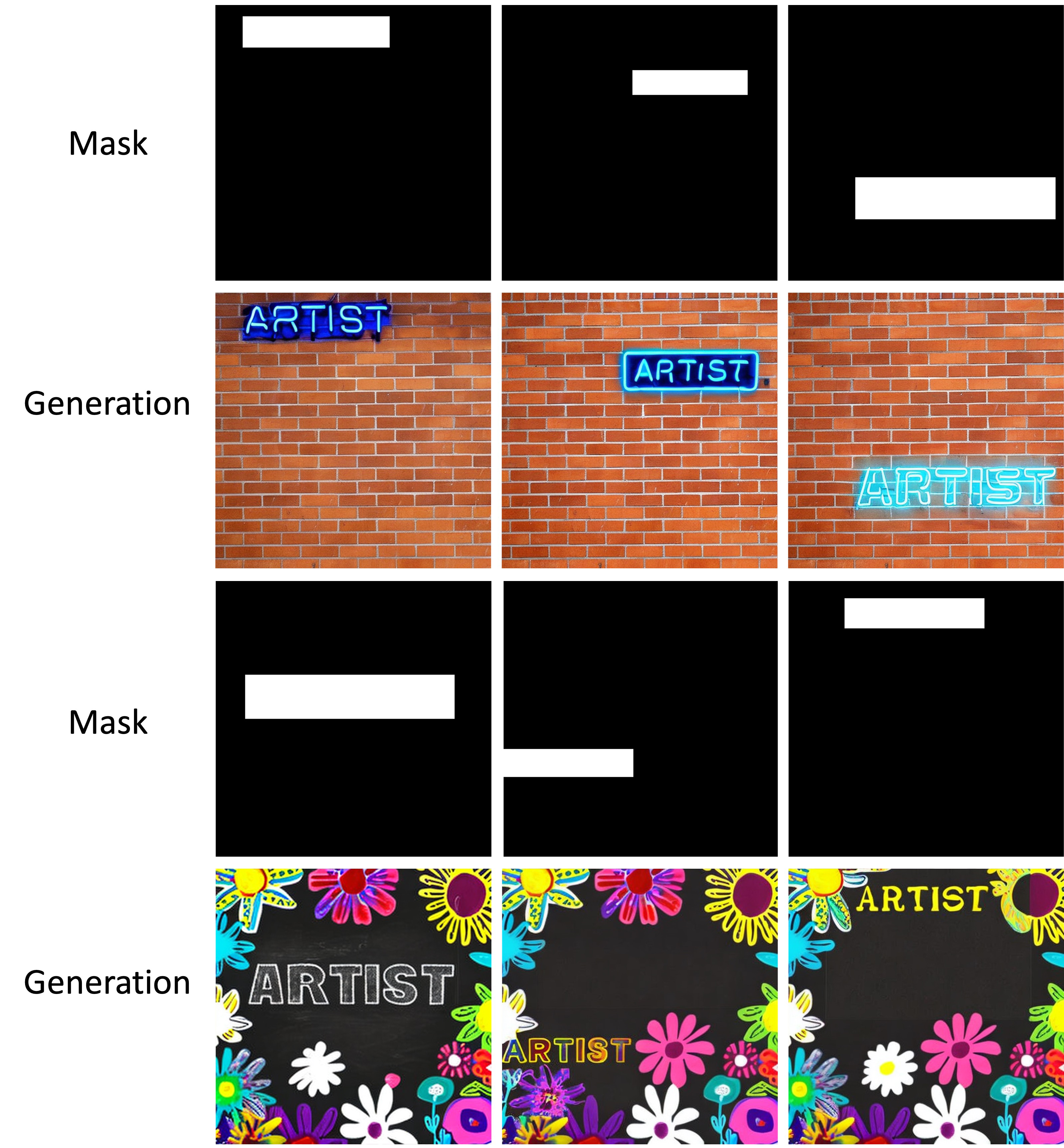}
    \caption{The generated text will follow the layout information in terms of mask positions and sizes. Prompts used here are ``a neon light `ARTIST' on the brick wall" and ``word `ARTIST' surrounded by hand-drawn flowers".}
    \label{fig:mask_generation}
    \vspace{-2em}
\end{figure}

\vspace{-4mm}
\paragraph{Layout Adherence}
We conduct an ablation study to validate the precision of our model in adhering to the provided layout during inference. Illustrative examples depicted in Figure \ref{fig:mask_generation} demonstrate that the text generated by our method consistently aligns with the specified position and dimensions. This consistency underscores the model's capability to follow layout guidelines accurately.

\section{Conclusions}
We proposed ARTIST, a novel framework that significantly enhances the text-rendering ability of diffusion models. Our proposed framework utilizes pretrained large language models to infer the user's intention, provide accurate prompts, and improve the interactive experience. We also introduced a disentangled architecture design and training strategy which leads to better learning of both text structure and visual appearance. Our experimental results demonstrate that ARTIST outperforms the previous state-of-the-art in terms of image fidelity, image-prompt alignment, and accuracy of generated texts. 
In the future, we aim to improve the disentangled learning further, investigate the interpretability of the disentangled representations learned by ARTIST
and their potential for downstream tasks. Overall, we believe that ARTIST represents a significant step forward in the field of text-rich image generation and has the potential to enable new applications in the future.

{\small
\bibliographystyle{ieee_fullname}
\bibliography{egbib}
}

\newpage
\appendix
\onecolumn

\section{ARTIST-Eval benchmark}\label{app:artist_benchmark}
\begin{figure}[htp]
    \centering
    \includegraphics[width=0.8\linewidth]{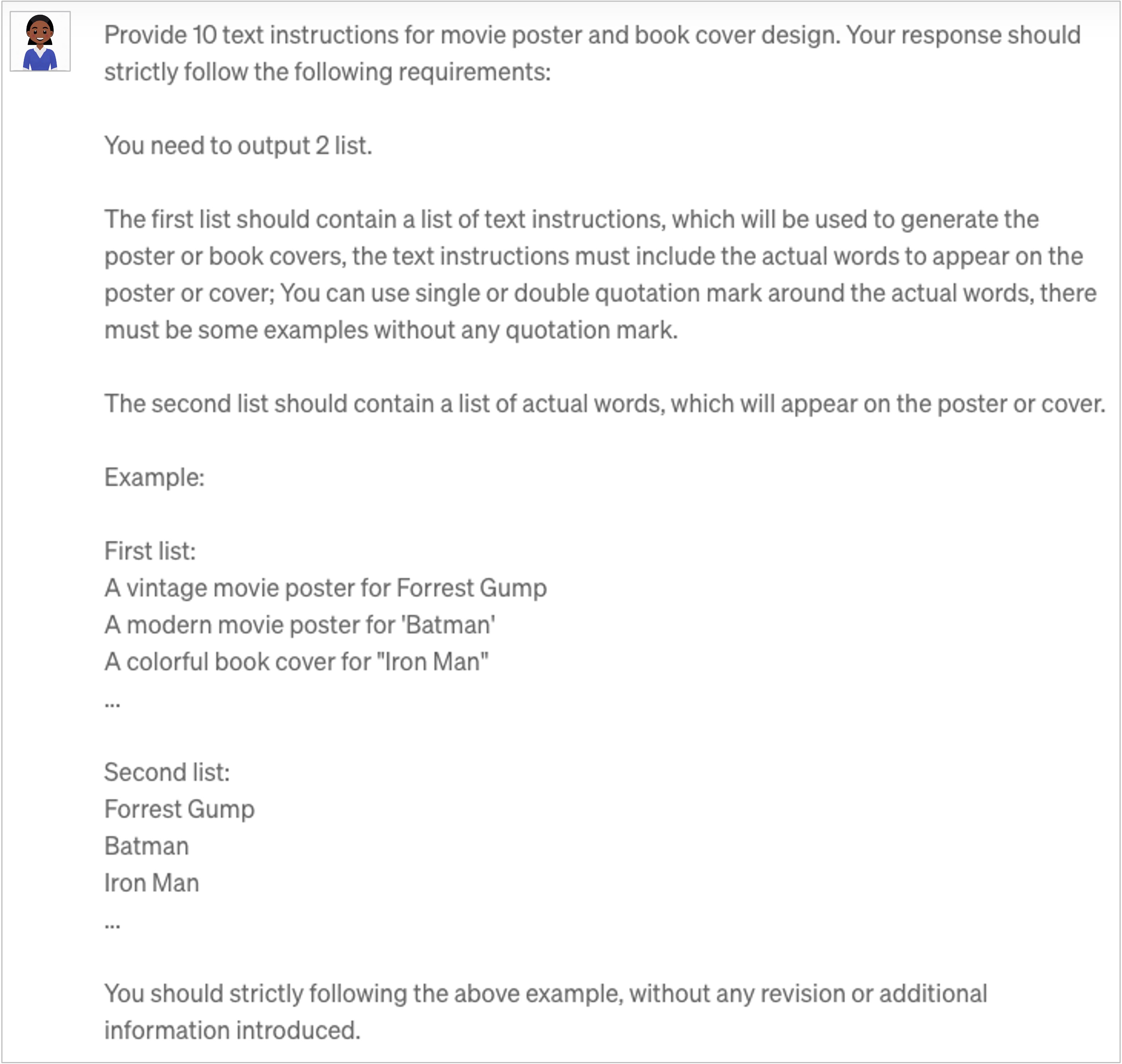}
    \caption{Prompt we used to construct our benchmark with GPT-4.}
    \label{fig:benchmark_prompt}
\end{figure}

Here we provide some examples in our ARTIST-Eval benchmark:
\begin{itemize}
    \item A vintage movie poster for Forrest Gump
    \item A modern movie poster for `Batman'
    \item A colorful book cover for ``Iron Man"
    \item A minimalist movie poster for The Godfather
    \item An abstract movie poster for `Pulp Fiction'
    \item A gothic book cover for ``Dracula"
    \item A romantic movie poster for The Notebook
    \item A futuristic movie poster for `Blade Runner'
    \item A watercolor book cover for ``Pride and Prejudice"
    \item A playful movie poster for Finding Nemo
\end{itemize}
Our benchmark is constructed by prompting GPT-4~\cite{OpenAI2023GPT4TR}, the prompt we used is shown in Figure \ref{fig:benchmark_prompt}.

\section{More Results}

\begin{figure}[h!]
    \centering
    \includegraphics[width=1.0\linewidth]{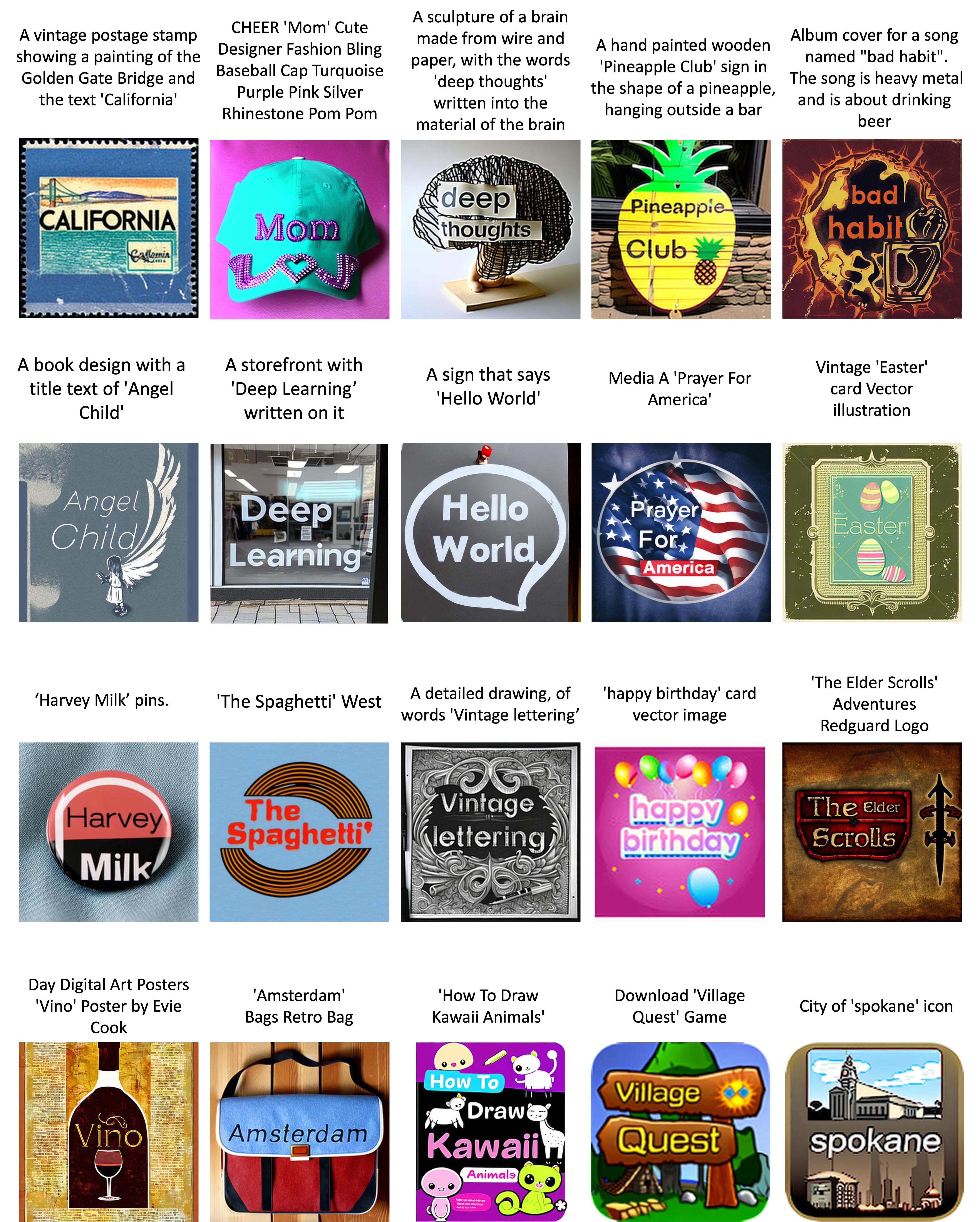}
    \caption{More generated results from the proposed framework.}
    \label{fig:more_results}
\end{figure}

\section{Examples from the survey}
\begin{figure}[h!]
    \centering
    \includegraphics[width=1.0\linewidth]{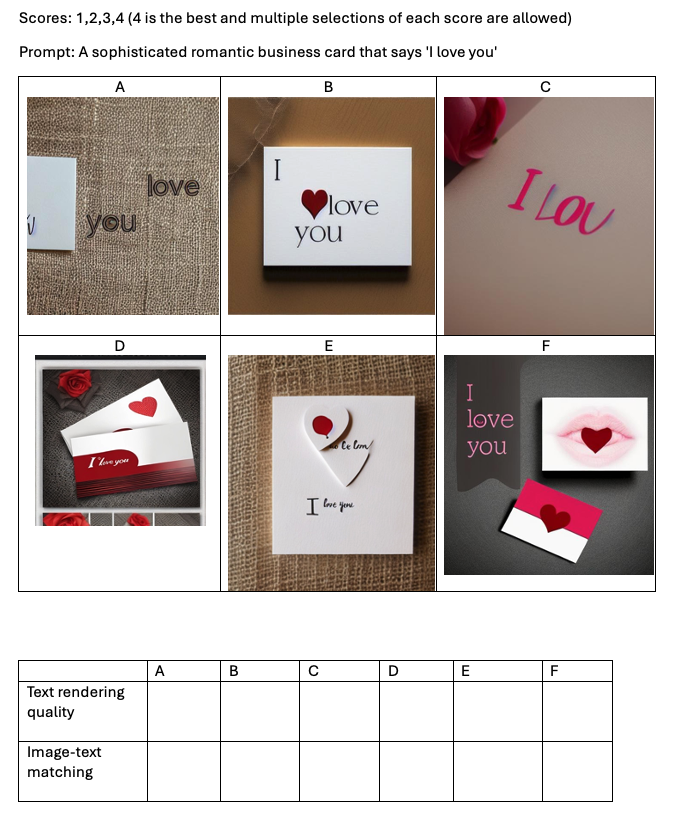}
    \caption{Example 1}
\end{figure}

\begin{figure}[h!]
    \centering
    \includegraphics[width=1.0\linewidth]{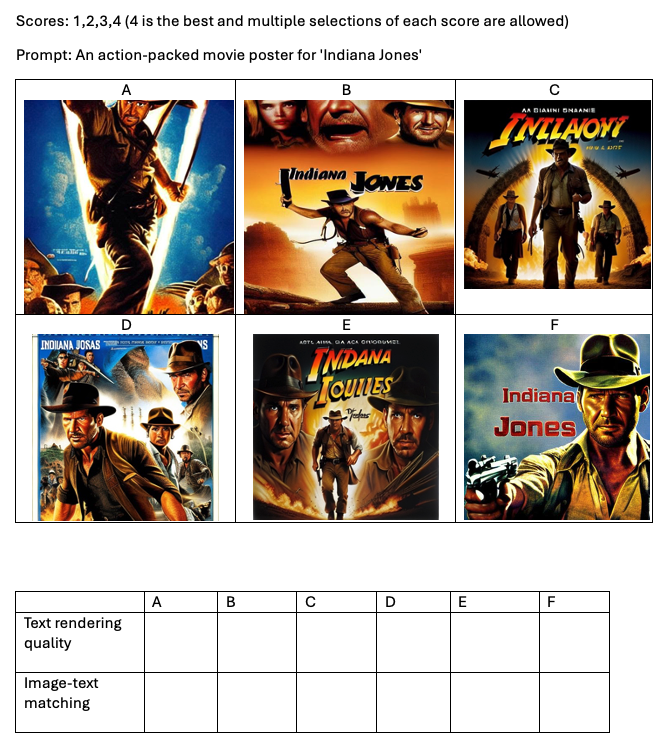}
    \caption{Example 2}
\end{figure}

\end{document}